\documentclass[letterpaper]{article} % DO NOT CHANGE THIS
\usepackage{aaai24}  % DO NOT CHANGE THIS
\usepackage{times}  % DO NOT CHANGE THIS
\usepackage{helvet}  % DO NOT CHANGE THIS
\usepackage{courier}  % DO NOT CHANGE THIS
\usepackage[hyphens]{url}  % DO NOT CHANGE THIS
\usepackage{graphicx} % DO NOT CHANGE THIS
\usepackage{natbib}  % DO NOT CHANGE THIS AND DO NOT ADD ANY OPTIONS TO IT
\usepackage{caption} % DO NOT CHANGE THIS AND DO NOT ADD ANY OPTIONS TO IT
\usepackage{algorithm}
\usepackage{algorithmic}
\usepackage{float}
\usepackage{cite}
\usepackage{graphicx}
\usepackage{makecell} 
\usepackage{multirow}
\usepackage[table,xcdraw]{xcolor}
\usepackage{tabularx}
\newcolumntype{"}{@{\hskip\tabcolsep\vrule width 2pt\hskip\tabcolsep}}
\usepackage{newfloat}
\usepackage{dirtytalk}
\usepackage{listings}
\usepackage{newfloat}
\usepackage{listings}
\DeclareCaptionStyle{ruled}{labelfont=normalfont,labelsep=colon,strut=off} % DO NOT CHANGE THIS
\floatstyle{ruled}
\newfloat{listing}{tb}{lst}{}
\floatname{listing}{Listing}
\title{FakeWatch ElectionShield: A Benchmarking Framework to Detect Fake News for Credible US Elections}
\author{
    Tahniat Khan\textsuperscript{\rm 1},
    Mizanur Rahman\textsuperscript{\rm 2},
    Veronica Chatrath\textsuperscript{\rm 1}, 
    Oluwanifemi Bamgbose\textsuperscript{\rm 1}, 
    Shaina Raza\textsuperscript{\rm 1}\\
}
\affiliations{
    \textsuperscript{\rm 1}Vector Institute, Toronto, ON, M5G 1M1, Canada\\
    \{tahniat.khan,veronica.chatrath, oluwanifemi.bamgbose,shaina.raza\}@vectorinstitute.ai \\
    \textsuperscript{\rm 2} Royal Bank of Canada, ON, Canada  \\
    mizanur.york@gmail.com
}

\usepackage{bibentry}
\nocopyright
\begin{document}

\maketitle

\begin{abstract}
In today's technologically driven world, the spread of fake news, particularly during crucial events such as elections, presents an increasing challenge to the integrity of information. To address this challenge, we introduce \textit{FakeWatch ElectionShield}, an innovative framework carefully designed to detect fake news. We have created a novel dataset of North American election-related news articles through a blend of advanced language models (LMs) and thorough human verification, for precision and relevance. We propose a model hub of LMs for identifying fake news. Our goal is to provide the research community with adaptable and accurate classification models in recognizing the dynamic nature of misinformation. Extensive evaluation of fake news classifiers on our dataset and a benchmark dataset shows our that while state-of-the-art LMs slightly outperform the traditional ML models, classical models are still competitive with their balance of accuracy, explainability, and computational efficiency. This research sets the foundation for future studies to address misinformation related to elections. 
\end{abstract}

\section{Introduction}

Fake news encompasses false or misleading information presented as if it were true, with the intent to deceive or manipulate. This deceptive practice is achieved through misinformation, which is spread without deceptive intent, and disinformation, which is deliberately created to mislead \cite{zhou_survey_2020,raza_fake_2022}. Fake news spreads through channels including traditional media, social media, websites, and online platforms, encompassing fabricated stories, distorted facts, sensationalized headlines, and selectively edited content. The motivation for creating and spreading misinformation varies from financial gain, to advancing a specific agenda, to acting as a tool for propaganda and influencing public opinion, to sowing confusion.

The negative consequences of fake news are evident in real-world scenarios. During the 2022 Ukraine-Russia conflict, termed \say{World War Wired} for its heavy social media documentation, fake news has been a major issue \cite{wright_portrayals_2023}. Fake news is widespread on online platforms, as highlighted by a NewsGuard report, showing that new TikTok users quickly encounter misleading war content \cite{russiaUkraine}. With the ongoing conflict, showing truth from falsehood remains a challenge for citizens, exacerbated by social media platforms' struggle to control fake news. 

During the 2020 Munich Security Conference, the Director-General of the World Health Organization (WHO) remarked, \say{We are not just fighting an epidemic; we are fighting an infodemic}, underscoring the alarming speed at which COVID-19 misinformation was spreading, compared to the virus itself \cite{muhammed_t_disaster_2022}. Over 6,000 people were hospitalized in the first three months of 2020 due to COVID-19 misinformation, with false claims of microchip implants contributing to vaccine hesitancy \cite{VaccineMyths}.

Fake news is not limited to these aforementioned specific events; it also impacts various aspects of society. Whether during elections, public health crises, or conflicts between nations, combating fake news has become crucial, emphasizing the role of artificial intelligence (AI) classifiers in identifying and mitigating misinformation. Integrating advanced technology and collaborative efforts is essential for navigating the information landscape and fostering a resilient, informed society.

Our research in fake news detection advances existing transformer-based model studies, particularly those by \cite{raza_fake_2022} and \cite{alghamdi_towards_2023}, with a focus on the dynamic context of North American elections. Previous insights from \citet{kaliyar_fakebert_2021, aimeur_fake_2023, hamed_review_2023} highlight deep learning techniques, but reveal shortcomings in handling data and concept drift \cite{raza2019news}. Data drift, the performance decline of AI models with new data, and concept drift, the evolution of data patterns, pose challenges in maintaining classifier accuracy, especially critical in the rapidly changing arena of election news. To solve these challenges, our research focuses on the 2024 US elections, tailoring our data preparation methods to address data and concept drift. 

We present three significant contributions that form the core of our work:

\begin{enumerate}

    \item \textbf{Dataset Compilation and Labeling} We have introduced a novel 2024 US Elections dataset, curated using targeted keywords and themes, and annotated through a combination of large language models (LLMs) and human verification. Existing datasets primarily focus on US Elections in 2016 and 2022 \cite{allcott2017social,grinberg2019fake,gruppi_nela-gt-2022_2023}. These datasets may not adequately address potential shifts in data and concepts during the upcoming 2024 US Elections. Hence, our new dataset is designed to address this gap for making robust fake news classifiers for the 2024 Presidential US Elections. Data prepared for this paper is available here \footnote{Anonymous due to double blind policy.}.
    
    \item \textbf{Comprehensive Hub of Classifiers as Benchmarks} We present an extensive range of machine learning (ML) and deep learning (DL) classifiers for fake news, creating a versatile and comprehensive hub. We have designed this model hub to provide invaluable and adaptable resources in the fight against misinformation.

    \item \textbf{Pioneering an Election-Specific Work} 
    To the best of our knowledge, we are the first to create a unique dataset and fake news classifier for the upcoming \textit{2024 US elections}, using AI to counter misinformation in this domain. Our dataset is specifically designed to address concept and data drifts, ensuring accurate analysis in the fast-changing election environment.
    
\end{enumerate}

\section{Related Works}
Fake news detection is a subtask of text classification \cite{liu_two-stage_2019}, commonly defined as the task of classifying news as either real or false. The term \say{fake news} refers to false or misleading information presented as authentic, with the intention to deceive or mislead people \cite{lu_fake_2022,zhou_survey_2020,allcott2017social,raza_fake_2022}. Fake news takes various forms, including clickbait (misleading headlines), disinformation (intentionally misleading the public), misinformation (false information regardless of motive), hoax, parody, satire, rumor, deceptive news, and other forms, as discussed in \citet{zhou_survey_2020}.

The recent developments in fake news classification highlight the integration of ML and DL techniques to improve accuracy and efficiency. A recent deep learning-focused study by \citet{lu_fake_2022} shows how these methods can enhance classifier performance. A related work by \citet{arora_reviewing_2023} provides a critical analysis of various algorithms used for fake news classification, shedding light on their effectiveness. Another work by \citet{bonny_detecting_2022} demonstrates the potential of ML classifiers in identifying fake news within English news datasets. In a similar work, a model is trained on detecting political news and fake information \cite{raza_automatic_2021}. These studies represent the state-of-the-art efforts in combating misinformation with AI. However, a key challenge in building an AI-based fake news classifier is related to datasets, feature representation, and data fusion.  \cite{hamed_review_2023}.

\textbf{Datasets}: The quality and diversity of datasets used for training and evaluating fake news detection models is a key challenge. Ensuring datasets are representative is crucial for models to generalize to real-world scenarios \cite{raza2023constructing}.  A comprehensive and balanced dataset is required to capture the multifaceted nature of fake news \cite{pubmed}. Benchmark datasets like Fakenewsnet \cite{shu2020fakenewsnet}, Fakeddit \cite{nakamura2019r}, NELA-GT \cite{gruppi_nela-gt-2022_2023} with different versions used for fake news detection

In response to the issue of COVID-19-related misinformation, \citet{alghamdi_towards_2023} implemented transformer-based models for fake news detection. Elections and fake news have always been a topic of interest, and many datasets have been proposed for this purpose \cite{allcott2017social,grinberg2019fake,gruppi_nela-gt-2022_2023}. These datasets are from the 2016 and 2022 US Elections and may not be able to combat data and concept drifts in the 2024 US Elections, which led us to develop a new dataset along this line.
  
\textbf{Feature Representations}: There are a number of challenges related to selecting relevant features, considering the dynamic nature of fake news content such as the temporal evolution of linguistic patterns \cite{raza2019news}, the influence of multimedia elements \cite{qi2019exploiting}, and the contextual nuances that contribute to more accurate feature representation \cite{fakeNewsData}.  \cite{raza_fake_2022,fakeNewsData} recognize the difficulty in identifying fake news due to the interplay between content and social factors, and propose a transformer-based approach, focusing on news content and social contexts.

\textbf{Data Fusion}: The integration of multiple data sources, or data fusion, is another critical challenge. Combining information from various modalities, such as textual content \cite{fakeNewsData}, user metadata \cite{raza_fake_2022}, and network structures \cite{aimeur_fake_2023}, and using LMs \cite{kaliyar_fakebert_2021} can make fake news detection systems more robust. Our work draws inspiration from prior research, with a particular emphasis on data and feature representations. Our novel approach of curating fresh election agenda data and uniquely applying LMs and ML techniques.

\begin{figure*}[h!]
    \centering
    \includegraphics[width=0.34\linewidth, angle=-90]{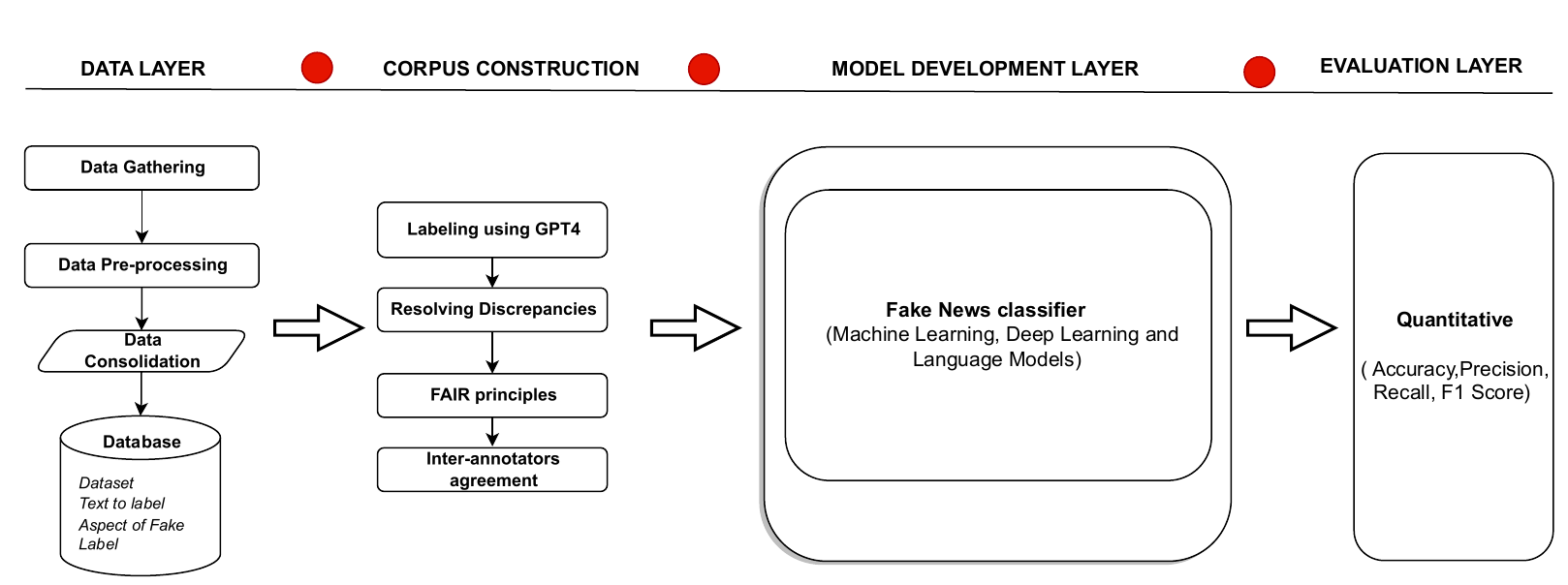}
    \caption{We present \textit{FakeWatch ElectionShield}, a framework to detect biases within textual data. In our proposed four-layer framework, data is first gathered from diverse sources and is then constructed into a quality-focused corpus. The data is processed by various machine learning models, and evaluated based on metrics such as accuracy, precision, recall, and F1 Score.}
    \label{fig:fig1}
\end{figure*}

\section{Framework for Fake News Identification}

\subsection{Defining Fake News}
Fake news detection can be defined as a binary classification problem, where each news item is assigned a label indicating its correctness. Mathematically, this can be represented as:

Let $N$ be a set of news items, where each news item, $n \in N$, is represented by a feature vector, $x_n$, containing its characteristics (e.g., text, metadata, etc.).
There is a corresponding label, \(y_{n}\), for each news item, \emph{n}, where \(y_{n} \in \{ 0,1\}\).

\(y_{n} = 1\) means the news item is fake (false information).

\(y_{n} = 0\) means the news item is real (true information).

A classification model, \emph{f}, is trained to map the feature vector, \(x_{n}\), to its corresponding label, \(y_{n}\) (i.e., vector
\({f(x}_{n}) \rightarrow y_{n}\)). The goal of this model is to
accurately predict the label for each news item based on its features,
thus distinguishing between fake and real news.

\subsection{FakeWatch ElectionShield}
\label{fakewatch-electionshield}

We present \textit{FakeWatch ElectionShield}, an innovative framework designed to detect biases within textual data, as illustrated in Figure \ref{fig:fig1}. The FakeWatch ElectionShield framework is structured into four distinct layers: (i) the data collection layer, (ii) the corpus construction layer, (iii) the model development layer, and (iv) the evaluation layer. Each layer is designed to integrate with the others, providing an effective and comprehensive approach for detecting fake news in text.

\subsubsection{Data Layer}

In this layer  we integrate data from two distinct sources: 1) Google RSS for data curation and 2) NELA-GT-2022 dataset \cite{gruppi_nela-gt-2022_2023}, an existing benchmark.

\textit{Data Curation:} We curated data from Google RSS by carefully selecting keywords, categorized into groups such as race/ethnicity-related terms, religious terms, geographical references, historical and political events, and other terms associated with racial discourse. From the Google RSS feeds, we use web scraping and news feed analysis to gather and categorize a wide array of news data from the US over a six-month period (Apr. 20, 2023 - Oct. 20, 2023). From the \textit{newspaper} library, we parse a selection of articles from each feed, focusing on: the first five sentences, the source of publication, publication date, and the article's URL. We then systematically categorize each article based on its corresponding keyword group and store it in a structured format.

After gathering and pre-processing data from the sources, we consolidate all datasets into a single data frame, resulting in a comprehensive dataset consisting of 9000 records, prioritizing quality over quantity. To safeguard users' privacy, we do not collect user IDs, and employ tokenization to replace references to usernames, URLs, and emails, ensuring all personally identifiable information is private. This dataset will be shared on Hugging Face to support the wider research community and facilitate further studies in this area.

The NELA-GT-2022 dataset is a benchmark dataset that provides source-level labels for each news article. We initially used these labels from NELA-GT-2022 to consolidate the dataframes (with curated data). From the NELA-GT-2022 data we filtered 5000 records in chronological order from Oct. 2022 - Dec.2022.

\textit{Data Consolidation:} We consolidate data from both sources (curated and NELA-GT) which contains the following columns:
\begin{itemize}
    \item \textit{Dataset:} Specifies the source dataset (e.g., news sources like BBC, CNN, etc.),
    \item \textit{Text: }Contains the actual textual data extracted from the respective datasets,
    \item \textit{Label:} Indicates whether the text is Fake (1) or Real (0), serving as the target variable for the token classifier and for evaluation purposes.
\end{itemize}

In the consolidated dataframe, each row represents a unique sample from the original dataset, supplying information for fake news detection and assessment. Further pre-processing is conducted to prepare the data for subsequent layers of the framework, particularly the natural language processing (NLP) model, performing token classification.

We improved data for ML algorithms by tokenizing and handling missing data. Tokenization breaks text into meaningful units, aiding semantic understanding. Managing missing data prevents bias and boosts model performance. These preprocessing steps create structured data for the NLP token classification model.

\subsubsection{Corpus Construction}
\label{corpus-construction}

\textit{Data Labelling:} The NELA-GT-2022 dataset includes source-level labels (e.g., BBC, CNN, The Onion), reflecting potential biases associated with news sources. Therefore, it is essential to annotate each individual news article. Similarly, the Google RSS curated data lacks labels for fake news detection, necessitating a strategy for news article annotation.

We employed OpenAI's GPT-4 \cite{gpt4} for initial labelling, assessing the likelihood of each news item being fake or real based on its language understanding capabilities. The use of data annotation through GPT-based models is also supported in the literature \cite{gilardi2023chatgpt} recently.

\textit{Data Quality:} Recognizing the importance of human expertise, we engaged six experts, including ML Scientists, Data Scientists, Linguistic Experts, and students, for manual verification of all 9,000 records. This comprehensive process validated and refined the AI-generated labels, capturing nuances that might have been overlooked by the automated system. Our hybrid approach, blending AI efficiency with human scrutiny, significantly enhanced the dataset's robustness.

Strict protocols were implemented to ensure reliable and consistent data labeling. Each record was independently reviewed by two experts, with their agreement measured using Cohen’s Kappa coefficient. A score of 0.80 in this metric, considered "almost perfect," was achieved, confirming the high uniformity and quality of our annotations and reinforcing the credibility of our dataset.

\subsubsection{Model Development Layer}

In this model development layer, we establish a comprehensive hub for fake news classification, encompassing traditional ML algorithms and advanced transformer-based models, including LMs. Our objective is to showcase the strengths of these diverse approaches, improving the accuracy and efficiency of fake news detection. We aim to deliver a robust and adaptable solution to combat misinformation, offering valuable insights into the real-world performance and scalability of these methods.

\begin{figure}[h!]
    \centering
    \includegraphics[width=1\linewidth]{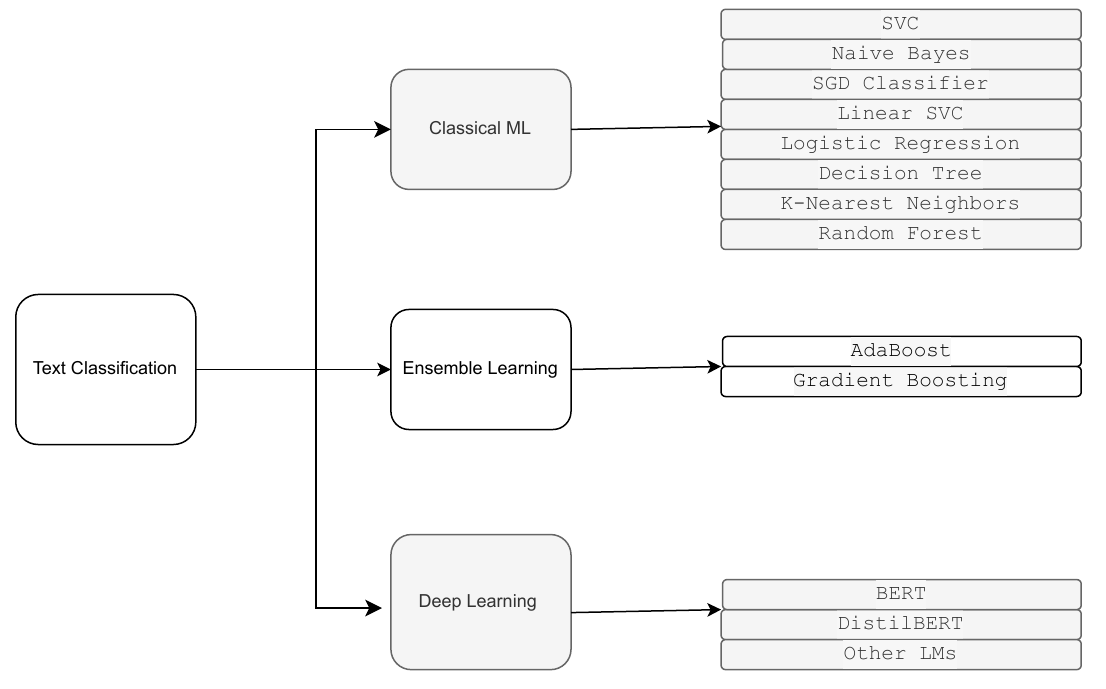}
    \caption{Classification Methods}
    \label{fig:classificationModels}
\end{figure}
To facilitate a structured comparison, we categorize the chosen models into three distinct groups, as depicted in Figure \ref{fig:classificationModels}. The models considered for comparison, with detailed definitions available in Section \ref{model_information} of the Appendix, are strategically grouped as follows: Naive Bayes, Logistic Regression, SGD Classifier, Random Forest, SVC (Support Vector Classifier), Linear SVC, and Decision Tree fall under the classical machine learning category. The ensemble learning group comprises Adaboost and Gradient Boosting. In the realm of deep learning, we include DistilBERT and BERT (Bidirectional Encoder Representations from Transformers).

\subsubsection{Evaluation layer}

The evaluation layer plays a critical role in assessing the performance of our model, using quantitative evaluation methods to provide a comprehensive breakdown on the model’s performance. We use measures, such as  F1 score, precision (the model's accuracy in identifying positive instances), and recall (the model's ability to recognize all relevant instances) to quantify the model's performance numerically.

\section{Experiments}
\label{experiments}
  
\subsubsection{Dataset and Exploratory Analysis} Our experiments are performed using our consolidated corpus of curated NELA-GT-2022 and Google RSS data, labelled with GPT-4 followed by comprehensive manual verification, as explain in Section \ref{fakewatch-electionshield}. 

We use an 80-20 train-test split on the dataset for all models. To address data imbalance, we employ an upscaling technique to ensure parity between different classes in the training set.

We conduct an exploratory analysis of the consolidated data and here we present the main findings, for brevity.

\begin{figure}[h!]
    \centering
    \includegraphics[width=1\linewidth]{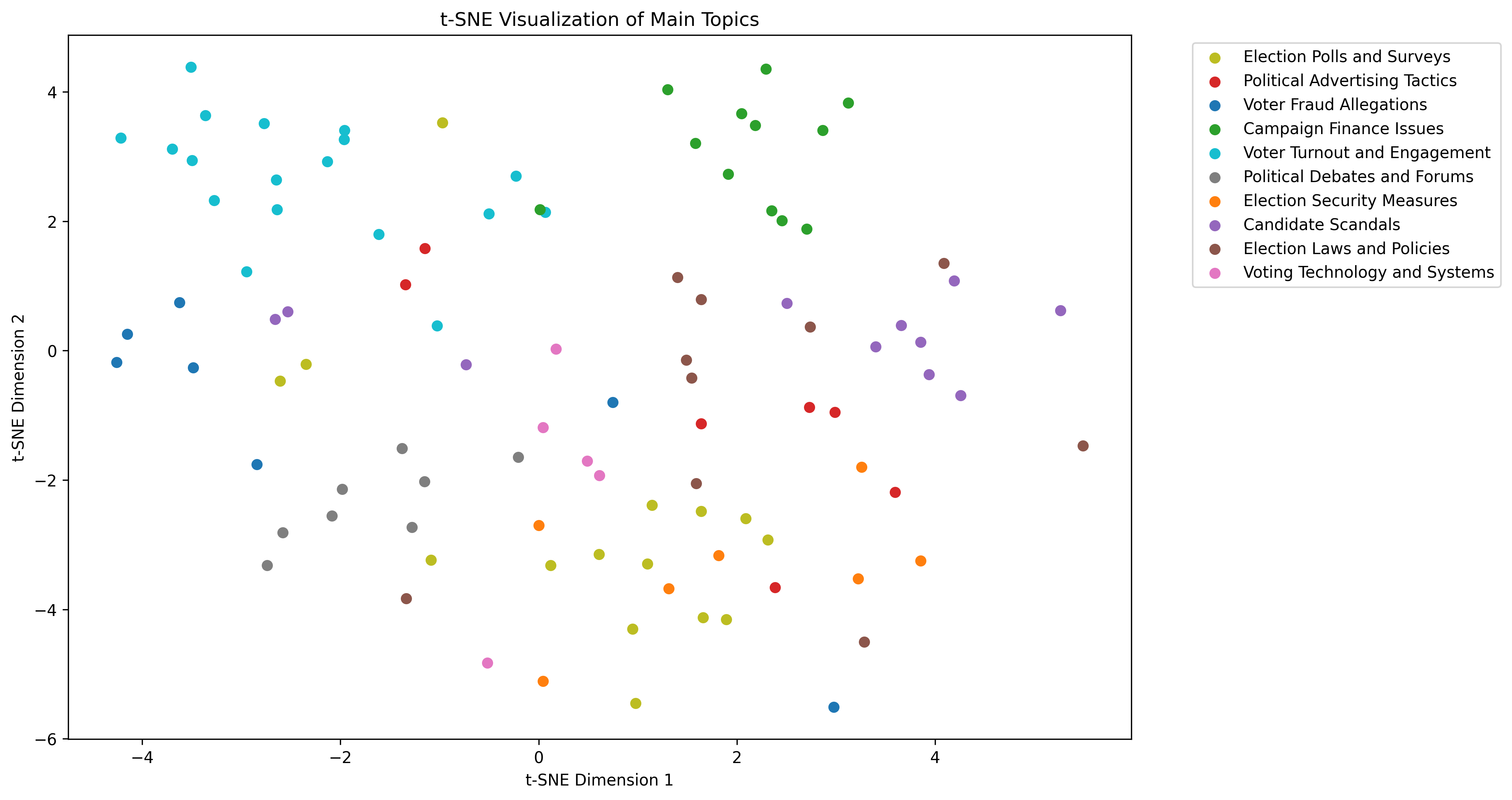}
    \caption{Important topics extracted from corpus.  Each point represents a document, and the color of the point indicates its most dominant topic, labelled according to the legend. Similar content clusters are based on dominant topics, and different topics are positioned farther apart.}
    \label{fig:topicsVis}
\end{figure}

\begin{figure}[h!]
    \centering
    \includegraphics[width=1\linewidth]{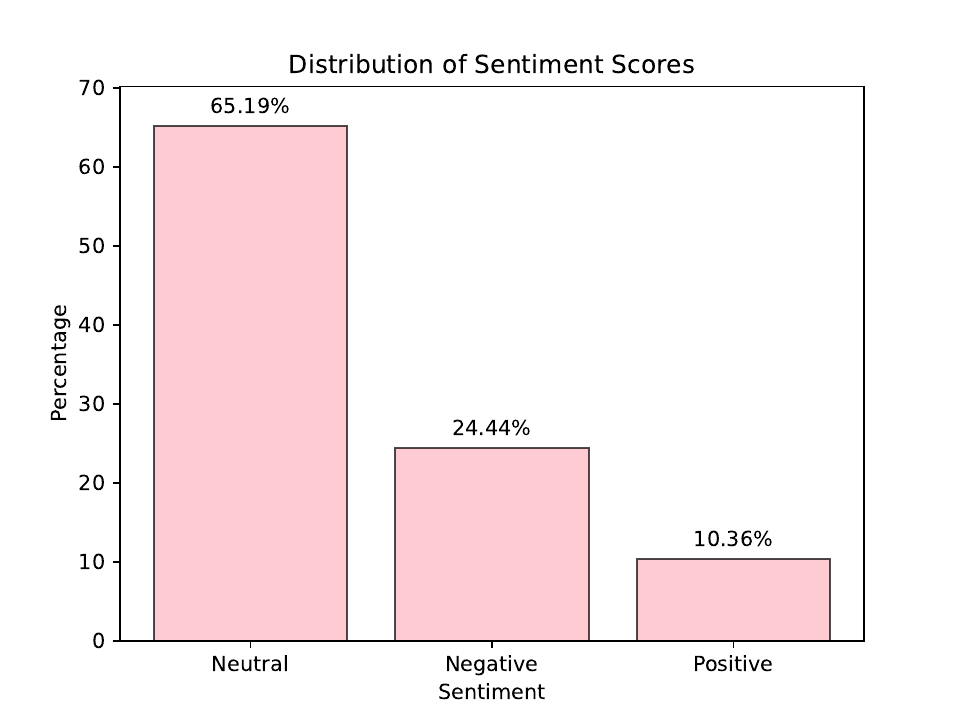}
    \caption{Scores for Neutral, Negative, and Positive sentiments.}
    \label{fig:sentiments}
\end{figure}

% \begin{figure}[h!]
%     \centering
%     \includegraphics[width=1\linewidth]{AuthorKit24/CameraReady/LaTeX/terms.png}
%     \caption{The most present terms in fake news datasets.}
%     \label{fig:presentTerms}
% \end{figure}

In Figure \ref{fig:topicsVis}, we use t-SNE (t-Distributed Stochastic Neighbor Embedding) on a subset of our data associated with 30 different election-related topics. Figure \ref{fig:topicsVis} shows that topics such as elections, political, votes and campaigns are closely placed.

We perform sentiment analysis on our corpus using pre-trained BERT for sequence classification using BERTopic \cite{grootendorst2022bertopic}. Figure \ref{fig:sentiments} shows a prevalence of neutral tones and a substantial amount of negative sentiments. The prevalence of neutral sentiments may indicate an effort to appear objective in disseminating misinformation, while the significant presence of negative sentiments suggests a strategy to evoke emotions like fear, distrust, or outrage, potentially amplifying the impact of fabricated narratives.

%%Furthermore, the scarcity of positive sentiments within the analyzed dataset hints at a potential trend where fake news articles are less inclined to convey affirmative or uplifting content. This pattern aligns with the common understanding that misinformation often relies on sensationalism, controversy, or alarmism to capture attention and manipulate public perceptions. Understanding the emotional nuances in fake news language is crucial for developing more effective detection models and devising strategies to counter the emotional manipulation embedded in misleading narratives.

\begin{figure}[h!]
    \centering
    \includegraphics[width=1.1\linewidth]{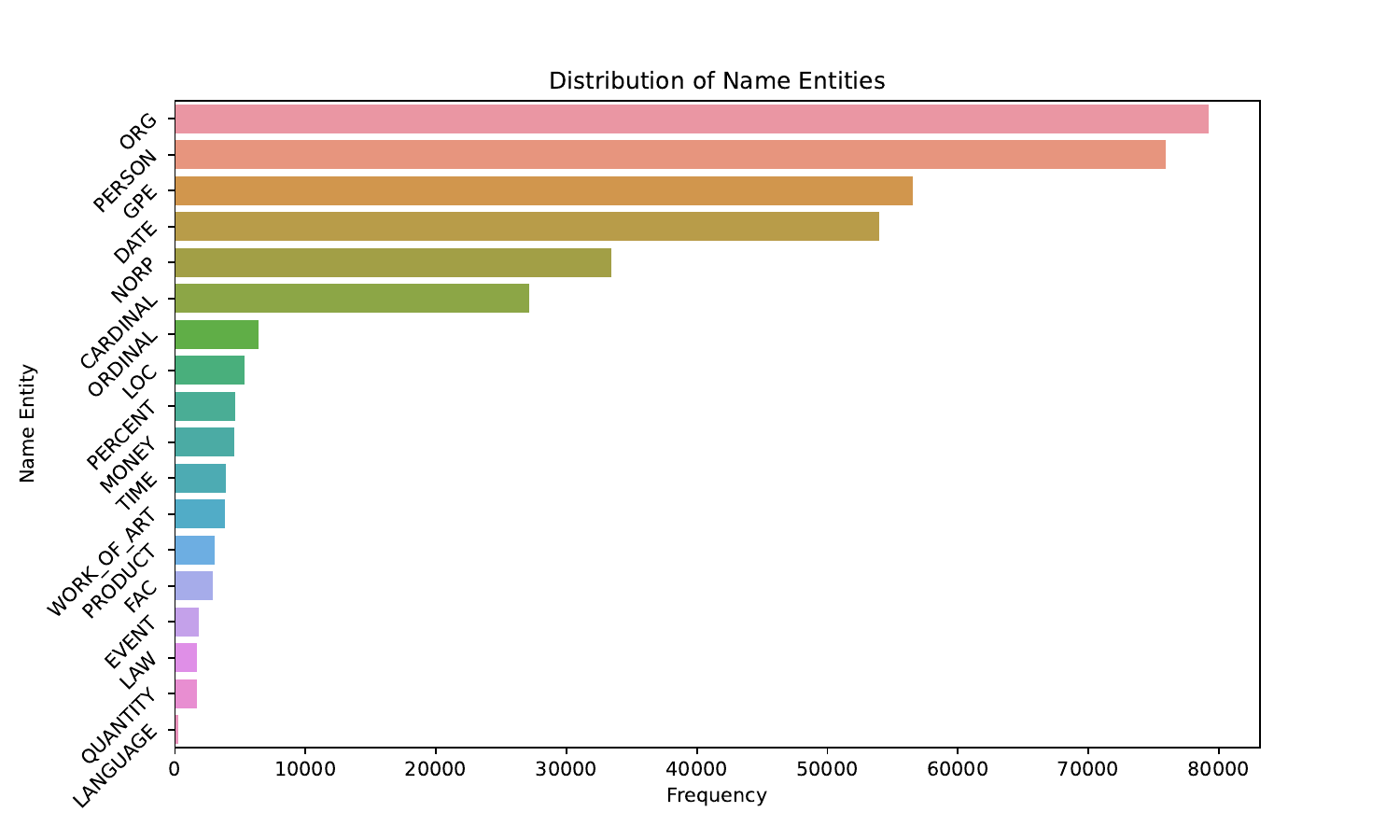}
    \caption{Distribution of Name Entities.}
    \label{fig:DistrNameEnt}
\end{figure}

Our NER analysis (Figure \ref{fig:DistrNameEnt}) on the data shows fake news predominantly targets specific organizations and individuals (ORG and PERSON) for sensational narratives. Additionally, it strategically focuses on nationalities, political groups, or religious communities using entities like GPE (Geopolitical Entity) to influence opinion, often accompanied by misleading dates (DATE). Numerical data is also manipulated in fake news to support false claims, highlighting the strategies used in its creation and dissemination.

\subsection{Evaluation Metrics}

\begin{table*}[h!]
\centering
\caption{Results of each model, sorted by F1 Score, with the highest values bolded.}
\label{results:tab1} 
\begin{tabular}{c|ccccc}
\hline
\Xhline{2\arrayrulewidth}  \Xhline{2\arrayrulewidth}
\textbf{Model} & \textbf{Accuracy $\uparrow$} & \textbf{Precision $\uparrow$} & \textbf{Recall $\uparrow$} & \textbf{F1 Score $\uparrow$} \\ \hline
\textbf{DistilBERT} & \textbf{0.80}     & \textbf{0.83} & \textbf{0.84}    & \textbf{0.84}                                                 \\ 
\textbf{BERT}                    & 0. 78                                                         & 0.81                                                           & 0.84                                                        & 0.83                                                          \\
\textbf{SGD Classifier}          & 0.79                                                          & 0.7                                                            & 0.49                                                        & 0.57                                                          \\
\textbf{Linear SVC}              & 0.78                                                          & 0.67                                                           & 0.49                                                        & 0.57                                                          \\
\textbf{Logistic Regression}     & 0.78                                                          & 0.69                                                           & 0.46                                                        & 0.56                                                          \\
\textbf{Bernoulli Naive Bayes}   & 0.66                                                          & 0.44                                                           & 0.68                                                        & 0.54                                                          \\
\textbf{SVC}                     & 0.78                                                          & 0.72                                                           & 0.42                                                        & 0.53                                                          \\
\textbf{Gradient Boosting}       & 0.77                                                          & 0.7                                                            & 0.35                                                        & 0.47                                                          \\
\textbf{Multinomial Naive Bayes} & 0.75                                                          & 0.67                                                           & 0.3                                                         & 0.42                                                          \\
\textbf{Decision Tree}           & 0.74                                  & 0.6                                                            & 0.32                                                        & 0.42                                                          \\
\textbf{AdaBoost}                & 0.75                                                          & 0.69                                                           & 0.3                                                         & 0.41                                                          \\
\textbf{K-Nearest Neighbors}     & 0.74                                                          & 0.62                                                           & 0.28                                                        & 0.4                                                           \\
\textbf{Random Forest}           & 0.74                                                          & 0.75                                                           & 0.19                                                        & 0.3                 \\     
\hline
\end{tabular}
\end{table*}

Accuracy, precision, recall, and F1 score are the evaluation metrics used for both classical ML models and transformer-based models. When comparing classical ML models with transformer-based models, these metrics help identify the strengths and weaknesses of each approach. Transformer-based models, such as those based on architectures like BERT or GPT, are often superior due to their ability to capture contextual relationships in language. However, these models may come with increased complexity and resource requirements. By considering accuracy, precision, recall, and F1 score, we can holistically evaluate the trade-offs and benefits of different models in the context of fake news detection, aiding in the selection of the most suitable model for a given application. True positives (TP) are the number of correctly identified instances, false positives (FP) are the number of incorrectly identified instances, true negatives (TN) are the number of instances correctly identified as fake news, and false negatives (FN) are the number of instances incorrectly identified as fake news.
 
$$ Accuracy  = \frac{\ TP + TN}{TP + TN + FP + FN} $$

$$ Precision = \frac{TP}{TP + FP} $$

$$ Recall = \frac{TP}{TP + FN} $$

$$ F1\ Score = 2\ \  \times \ \ \frac{Precision\ \  \times \ \ Recall}{Precision + Recall} $$ 

These equations provide a quantitative measure of various aspects of a classification model's performance. Accuracy gives an overall view, while precision, recall, and F1 score focus on the trade-offs between false positives and false negatives. These metrics collectively help assess the model's ability to correctly classify instances in tasks such as fake news detection, where both false positives and false negatives have significant consequences.
\subsection{Experimental Settings}

We use an Intel(R) Core(TM) i7-8565U CPU, Google Colab Pro with cloud-based GPUs, and Google Drive for storage. We implement BERT encoder layers using PyTorch BERT from Hugging Face. Additionally, human assessment is conducted to validate the effectiveness of the evaluation strategies. General hyperparameters can be found in Table \ref{table1} in the Appendix.

\section{Results}
\label{sec:results}
\subsection{Overall Performance}
As seen in Table \ref{results:tab1}, we see the results from different models. We observe that DistilBERT model stands out with an impressive accuracy of 0.80. Its precision, recall, and F1 score are consistent and high at 0.83, 0.84, and 0.84, respectively. This indicates the model's robust performance in correctly classifying both positive and negative instances of fake news.  We also observe that the BERT model performs well in precision (0.81) and recall (0.84), contributing to a commendable F1 score of 0.83.  Both these results indicate that LMs (deep neural networks and transformer-based models) are quite good in accuracy.

In a comparative analysis of ML models, the SGD classifier shows notable accuracy (0.79) but a lower F1 score (0.57), indicating a mismatch between overall accuracy and precision-recall balance. Linear SVC and Logistic Regression exhibit similar performances with F1 scores around 0.56-0.57 and an accuracy of 0.78. The Bernoulli Naive Bayes demonstrates moderate accuracy (0.66) and an F1 score of 0.54, suggesting decent recall capability. SVC achieves a balanced accuracy and precision (0.78 and 0.72, respectively) but has a lower F1 score (0.53) due to reduced recall. Gradient Boosting and Multinomial Naive Bayes show moderate accuracies around 0.74-0.77 but lower F1 scores (0.47 and 0.42), indicating challenges in achieving high recall. AdaBoost and K-Nearest Neighbors have F1 scores just above 0.4 with reasonable accuracy, suggesting average classification effectiveness. The Random Forest model, despite high accuracy (0.74) and precision (0.75), has the lowest F1 score (0.3) due to very low recall, highlighting a significant limitation in its performance.

Our evaluations highlight the diverse landscape of fake news classification models, revealing performance variations. The traditional ML models showcase strengths in specific metrics, while transformer-based models, especially DistilBERT and BERT, demonstrate a strong balance between precision and recall. 

The performance gains between the two sets of models (simple ML and LMs) is not evident, which may reflect that classic ML models can be used if computational resources are constrained. These findings underscore the importance of considering a spectrum of metrics for a holistic evaluation, providing valuable insights for researchers and practitioners seeking effective models for fake news classification in real-world scenarios.

\subsection{Confusion Matrix Results}

Here, we provide additional confusion matrix values, as we believe a detailed breakdown is crucial for identifying areas where a model succeeds or falls short. With this additional information, we can improve precision and recall, ultimately enhancing the overall effectiveness of fake news detection systems.

\begin{table}[h]
\centering
\caption{Confusion matrix results, with the highest values bolded. TP is True Positive, FN is False Negative, and FP is False Positive.}
\label{Confresults}
\begin{tabular}{c|cccc}
\hline
\Xhline{2\arrayrulewidth}  \Xhline{2\arrayrulewidth}
\textbf{Model} & \textbf{TP $\uparrow$} & \textbf{FN $\downarrow$} & \textbf{FP $\downarrow$} \\ \hline
\textbf{DistilBERT}                                        & \textbf{72\%}                                           & \textbf{8\%}                                            & \textbf{5\%}                                                                                      \\
\textbf{BERT}                                              & 71\%                                                    & 5\%                                                     & 6\%                                                                                \\
\textbf{SGD Classifier}                                    & 65\%                                                    & 6\%                                                     & 15\%                                                                \\
\textbf{Linear SVC}                                        & 64\%                                                    & 7\%                                                     & 15\%                                                    \\
\textbf{Logistic Regression}                               & 65\%                                                    & 6\%                                                     & 15\%                                                        \\
\textbf{Bernoulli Naïve Bayes}                             & 47\%                                                    & 25\%                                                    & 9\%                                                     \\
\textbf{SVC}                                               & 67\%                                                    & 5\%                                                     & 17\%                                                    \\
\textbf{Gradient Boosting}                                 & 67\%                                                    & 4\%                                                     & 19\%                                                    \\
\textbf{Multinomial Naïve Bayes}                           & 67\%                                                    & 4\%                                                     & 20\%                                                    \\
\textbf{Decision Tree}                                     & 65\%                            & 6\%                                                     & 19\%                                                   \\
\textbf{Ada Boosting}                                      & 67\%                                                    & 4\%                                                     & 20\%                                                    \\
\textbf{K-Nearest Neighbors}                               & 66\%                                                    & 5\%                                                     & 21\%                                                    \\
\textbf{Random Forest}                                     & 70\%                                                    & 2\%                                                     & 23\%                         \\                      
\hline
\end{tabular}
\end{table}

As seen in Table \ref{Confresults}, DistilBERT demonstrates a high TP rate of 72\%, and a comparatively low FN rate of 8\%, suggesting its effectiveness in correctly classifying genuine news. BERT exhibits a similar trend with a TP rate of 71\% and a FN rate of 5\%. These results suggest that LMs are high in performance.

On the other hand, Bernoulli Naïve Bayes displays a TP rate of 47\%, accompanied by a notable FN rate of 25\%. Notably, Random Forest achieves a remarkable TP rate of 70\% and a very low FN rate of 2\%, indicating its proficiency in identifying fake news instances. These insights of TP and FN rates further underscores the strengths of transformer-based models like DistilBERT and BERT in correctly identifying real news. The notable performance of Random Forest, especially in achieving a high TP rate and a remarkably low FP rate, suggests that traditional ML models can still offer competitive results in the realm of fake news detection. 

The selection of an appropriate model should consider the trade-offs between true positive and false negative rates based on the specific priorities and requirements of the task at hand.

\section{Discussion}
\label{discussion}
\subsection{Main Findings}

Our research provides insights into the classification of fake news, particularly in the context of elections. We observe a notable prevalence of fake news on online platforms, with distinctive spread patterns during political events and crises. While ML methods show varying degrees of effectiveness in classification, challenges persist due to the subtlety of misinformation and the evolving nature of fake news content \cite{raza_nbias_2024}. 

Our novel dataset addresses these challenges, addressing both data and concept drift, crucial in training classification models. These drifts are pronounced in election-related fake news due to the dynamic nature of political discourse and evolving strategies for creating and spreading fake news. 

Our comprehensive analysis of fake news detection models reveals that transformer models like DistilBERT and BERT excel in fake news detection, demonstrating high accuracy and reliability. Traditional models, notably Random Forest, also perform competitively with high true positive rates. The study highlights the nuanced landscape of model effectiveness, emphasizing the importance of considering specific task requirements when selecting an appropriate model for tackling the challenge of fake news.

\subsection{Practical and Theoretical Impacts}
This research has significant practical implications. It can help media organizations and the public in developing better tools and awareness for identifying fake news. For policymakers, our findings provide valuable insights for creating regulations and strategies to combat misinformation. Furthermore, this research contributes to the advancement of detection tools and technologies, enhancing information integrity across platforms.

Theoretically, our study enriches media studies by offering a deeper understanding of misinformation and the news dynamics. It advances computational linguistics, particularly in enhancing natural language processing algorithms for news classification \cite{raza_fake_2022}. Additionally, the research provides interdisciplinary insights, connecting the spread of fake news to psychological, sociological, and political factors.

In the era of LLMs, these models serve a dual role in the fake news landscape. Although they can be exploited to generate convincing fake news, raising ethical concerns, LLMs also offer advanced capabilities in detecting and flagging fake news \cite{bang2023multitask, huang2023survey}. This dual role highlights the importance of developing ethical frameworks for the use of LLMs in news-related applications.

\subsection{Enhancing Data Labelling with Language Models}

Our research is committed to mitigating biases introduced by LLM-based labeling, through a set of proposed human verification steps, as detailed in \citet{gilardi2023chatgpt}. To expand our labeling efforts in the future, we will employ crowdsourcing. Some of the key human verification steps we will implement are as follows:

\begin{enumerate}
    \item \textit{Random Sampling for Quality Control}: We will regularly select a subset of labels generated by LLMs and subject them to review by human verifiers. This random sampling approach serves as a robust quality control measure.
    \item \textit{Expert Review for Sensitive Topics:} Categories involving sensitive subjects such as race, gender, or political opinions will undergo expert review. This step ensures that the labels are equitable and devoid of bias. 
    \item \textit{Feedback Loop}: The inputs provided by human verifiers will be integrated in refining the LLM's labeling process. 
\end{enumerate}
Additionally, for our crowdsourcing efforts, we will focus on two essential aspects:

\begin{enumerate}
    \item \textit{Diversity in the Verification Team:} When assembling our verification team, we will prioritize diversity.
    \item \textit{Training and Guidelines:} Crowdsourced verifiers will undergo training and be provided with clear guidelines to ensure consistency and fairness in their assessments. This training will help them understand the nuances of the labeling task and maintain high-quality standards.
\end{enumerate}

\subsection{Research Gaps and Future Directions}

Our study identifies several research gaps. Areas such as specific linguistic and cultural contexts in fake news classification remain underexplored. Furthermore, current methods face limitations, including inherent biases and a lack of adaptability to various news contexts, necessitating further research.

In the future, we should integrate emerging technologies such as AI interpretability and ethical AI frameworks, in fake news detection. The future of fake news classification lies in cross-disciplinary research that merges technology, psychology, and media studies. Developing adaptive algorithms capable of evolving with changing news narratives is a crucial direction for future exploration. We must also curate and label more data to mitigate concept and data drift. Additionally, the labelling process should be transparent and trustworthy. 

Nevertheless, this research lays a foundation in fake news data construction and provides a hub for further investigation and development.

\section{Conclusion}

In conclusion, we present a detailed approach to detecting fake news, from data collection and dataset creation to the development of models. Our work focuses on overcoming key challenges in this field. We ensure our dataset is balanced, to reduce biases, guaranteeing representation that is fair and covers a wide range of views and topics. We have also been careful about potential biases in data gathering and have taken steps to address these. Furthermore, our methodology includes a variety of machine learning models. By tackling these issues, our research not only strengthens the detection of fake news but also adds to the ongoing conversation about ethical and unbiased AI practices in verifying information.

\bibliography{aaai24}

%%\newpage
\section*{Appendix}

\subsection{Hyperparameters}
The following is a table of the hyperparameters used for all models in our experiments.

\begin{table}[h!]
\centering
\caption{Table of Hyperparameters.}
\label{table1}
\begin{tabular}{|p{2cm}"p{5.5cm}|}
\hline
\textbf{Model} & \textbf{Hyperparameters} \\
\Xhline{2\arrayrulewidth}  \Xhline{2\arrayrulewidth}

\hline
Naive Bayes                            & Alpha: 1.0, Fit Prior: \texttt{True}
\\ &\\
Logistic Regression                    & C: 1.0, Solver: \texttt{lbfgs}, Penalty: \texttt{l2} \\ &\\
SGD Classifier                         & Loss Function: \texttt{hinge}, Penalty: \texttt{l2}, Learning Rate: 0.01  \\ &\\
Random Forest                          & Number of Trees: 100, Max Depth: \texttt{None}, Max Features: \texttt{auto}  \\ &\\
AdaBoost                               & Number of Estimators: 50, Learning Rate: 1.0     \\ &\\
Gradient Boosting                      & Learning Rate: 0.1, Number of Estimators: 100, Max Depth: 3   \\ &\\
SVC                                    & Kernel: \texttt{rbf}, C: 1.0, Gamma: \texttt{scale}  \\ &\\
Linear SVC                             & C: 1.0, Loss: \texttt{squared\_hinge}, Penalty: \texttt{l2} \\ &\\
Decision Tree                          & Max Depth: None, Min Samples Split: 2, Criterion: \texttt{gini}   \\ &\\
DISTILBERT                             & Learning Rate: 6$\times 10^{-6}$, Number of Epochs: 8, Batch Size: 32 \\ &\\
BERT                                   & Learning Rate: 6$\times 10^{-6}$, Number of Epochs: 8, Batch Size: 32  \\ &\\
\hline
\end{tabular}
\end{table}

\subsection{Models Implemented}
\label{model_information}
\textbf{Naive Bayes}: A probabilistic classifier based on
Bayes\textquotesingle{} theorem, assuming strong independence between
features. It's known for its simplicity and
effectiveness, especially in text classification tasks like spam
detection.

\textbf{Logistic Regression:} A fundamental statistical model that
predicts the probability of a binary outcome based on input features.
It's widely used for binary classification problems,
such as credit scoring and medical diagnosis.

\textbf{SGD Classifier:} Stands for Stochastic Gradient Descent
Classifier. It's a linear classifier (like SVM or
logistic regression) that uses gradient descent to optimize the loss
function. Ideal for large-scale and sparse ML problems.

\textbf{Random Forest:} An ensemble learning method that constructs
multiple decision trees at training time and outputs the mode of the
classes (classification) of the individual trees. It's
great for handling a large dataset with high dimensionality.

\textbf{AdaBoost:} Short for Adaptive Boosting, it's an
ensemble technique that combines multiple weak classifiers to create a
strong classifier. It's known for its effectiveness in
improving the accuracy of any given learning algorithm.

\textbf{Gradient Boosting:} Another ensemble technique that builds trees
in a sequential manner, where each tree tries to correct the errors made
by the previous one. It's used widely for both
regression and classification problems.

\textbf{SVC (Support Vector Classifier)}: Part of the SVM (Support
Vector Machine) family, it's used for classification
problems. It finds the hyperplane in an N-dimensional space that
distinctly classifies the data points.

\textbf{Linear SVC:} A version of SVC with a linear kernel.
It's similar to SVC with the parameter kernel set to
'linear\textquotesingle, but implemented in terms of
liblinear rather than libsvm, so it has more flexibility in the choice
of penalties and loss functions and should scale better to large numbers
of samples.

\textbf{Decision Tree:} A tree-like model of decisions.
It's a type of supervised learning algorithm (having a
pre-defined target variable) used in statistics, data mining, and
ML.

\textbf{DISTILBERT:} A smaller, faster, cheaper, and lighter version of
BERT. It distills the crucial information from BERT, retaining 97\% of
its language understanding capabilities but with less computational
cost.

\textbf{BERT (Bidirectional Encoder Representations from Transformers)}
(Devlin et al., 2018)\textbf{:} A transformer-based ML
technique for natural language processing pre-training.
It is designed to understand the context of a word in a
sentence, which significantly improves the state-of-the-art in sentence understanding.

Each of these models brings its unique strengths to the task of fake news detection, allowing for a comprehensive approach that enhances the robustness and accuracy of our classification system.

\textbf{Language Models:}
Our methodology is adaptable to various Language Models (LMs) or Large Language Models (LLMs) due to the clarity of our data construction guidelines and the comprehensive implementation details we provide. This extensibility enables the application of our approach to a wide range of language models beyond the specific one considered in this study.

\end{document}